%% file: acl2023.tex
\pdfoutput=1

\documentclass[11pt]{article}

\usepackage[]{ACL2023}

\usepackage{times}
\usepackage{latexsym}

\usepackage[T1]{fontenc}

\usepackage[utf8]{inputenc}

\usepackage{microtype}

\usepackage{inconsolata}

\usepackage{amsfonts}
\usepackage{amsmath}
\usepackage{subcaption}
\usepackage{graphicx}
\usepackage{pgfplots} 
\usepackage{scalefnt} 

\usepackage[misc]{ifsym}

\definecolor{color1}{RGB}{199,47,47}
\definecolor{color2}{RGB}{242,129,29}
\definecolor{color3}{RGB}{52,157,53}
\definecolor{color4}{RGB}{142,111,173}
\definecolor{darkred}{RGB}{180,0,0}
\definecolor{darkblue}{RGB}{0,100,205}

\newcommand{\darkred}[1]{\textcolor{darkred}{#1}}
\newcommand{\darkblue}[1]{\textcolor{darkblue}{#1}}

\usepackage{footmisc}
\usepackage{scrextend}
\usepackage{blindtext}

\deffootnote[1em]{1.5em}{1em}{%
\textsuperscript{\thefootnotemark}%
}
\DefineFNsymbols*{noasterisk}{
   {\TextOrMath \textdagger \dagger}%
   {\TextOrMath \textdaggerdbl \ddagger}%
   {\TextOrMath \textsection  \mathsection}%
   {\TextOrMath \textparagraph \mathparagraph}%
   {\TextOrMath \textbardbl \|}%
   {\TextOrMath {\textdagger\textdagger}{\dagger\dagger}}%
   {\TextOrMath {\textdaggerdbl\textdaggerdbl}{\ddagger\ddagger}}%
}
\setfnsymbol{noasterisk}

\title{Probing Multimodal Large Language Models for Global and Local Semantic Representations}

\author{
Mingxu Tao$^{\text{1}}$\thanks{This work was done when Mingxu Tao and Quzhe Huang were interns at Kuaishou Technology.}, Quzhe Huang$^{\text{1}}$\footnotemark[1], Kun Xu$^{\text{2}}$, Liwei Chen$^{\text{2}}$, \textbf{Yansong Feng$^{\text{1\Letter}}$, Dongyan Zhao$^{\text{1}}$}\\
 $^{\text{1}}$Peking University\ \ \ $^{\text{2}}$Kuaishou Technology \\
         \texttt{\{thomastao, huangquzhe, zhaodongyan\}@pku.edu.cn}\\
         \texttt{syxu828@gmail.com\ \ \ chenliwei03@kuaishou.com}\\
         \texttt{\Letter fengyansong@pku.edu.cn}\\}

\begin{document}
\maketitle
\begin{abstract}
The advancement of Multimodal Large Language Models~(MLLMs) has greatly accelerated the development of applications in understanding integrated texts and images. Recent works leverage image-caption datasets to train MLLMs, achieving state-of-the-art performance on image-to-text tasks. However, there are few studies exploring which layers of MLLMs make the most effort to the global image information, which plays vital roles in multimodal comprehension and generation. 
In this study, we find that the intermediate layers of models can encode more global semantic information, whose representation vectors perform better on visual-language entailment tasks, rather than the topmost layers. We further probe models regarding local semantic representations through object recognition tasks. We find that the topmost layers may excessively focus on local information, leading to a diminished ability to encode global information. Our code and data are released via \url{https://github.com/kobayashikanna01/probing_MLLM_rep}.
\end{abstract}

\section{Introduction}

Recently, Large Language Models~(LLMs) have achieved remarkable advancements in various natural language processing applications~\cite{touvron2023llama,openai2024gpt4}, 
owing to pre-training on massive text corpus. It becomes a popular topic nowadays to transfer the powerful capacity of LLMs to Multimodal Large Language Models~(MLLMs) through image-caption corpus~\cite{alayrac2022flamingo,li2023blip2}. These MLLMs show an impressive ability to handle multimodal tasks, including Image Captioning~(IC, \citealp{Plummer_2015_Flickr30k}) and Visual Question Answering~(VQA, \citealp{Goyal_2017_vqa}). However, existing research has predominantly focused on the ability of MLLMs to generate single tokens one by one,  
while lacking investigations about how their representation vectors can encode global multimodal information. 
In generation tasks like IC and VQA, when predicting the next token, the models may only need to focus on a local part of the image and a subsequence of the text to handle the task. But in tasks like image-text retrieval~\cite{xie2019visual}, the MLLMs should aim to encode the global semantic information of the entire image and text, when predicting whether they have correlation. 

In this work, we focus on understanding and uncovering how the global and local semantic information is encoded in the decoder-only MLLMs. To track the representing ability of each layer in MLLMs, we use probing study, a popular tool to investigate model interpretability~\cite{tenney2018what,jawahar-etal-2019-bert}. Previous probing studies of pure-text language models have explored the representing ability of models in various levels, from local to global semantics~\cite{liu-etal-2019-linguistic,talmor-2020-olympics}. However, to the best of our knowledge, existing works about vision-language models sorely focus on investigating the ability to represent local semantic information, for instance, from a lexical perspective~\cite{dahlgren-lindstrom-etal-2020-probing}. We also note that previous works~\cite{ma-etal-2022-probing,dai-etal-2023-plausible} mainly study the encoder-only or encoder-decoder models, with less than 1B parameters, such as CLIP~\cite{clip} and BLIP~\cite{li2022blip}. Our work investigates the representing ability of decoder-only MLLMs, from both global and local perspectives, thus bridging a gap in prior works.

Our main contributions in this paper are:

(1) We design an image-text entailment task to probe MLLM's ability to encode global cross-modal information and design a pair of prompts for object recognition to study local representation.

(2) We find that, when encoding global information, it is the intermediate layers rather than the topmost layers that perform the best.

(3) Through the probing study of local representations, we find the topmost layers may excessively focus on local information, leading to a diminished ability to encode global information.

(4) To the best of our knowledge, we are the first to find and discuss the potential shortcomings of decoder-only MLLMs in representing global semantic information. We hope our findings could encourage the community to explore ways to improve the pre-training process of MLLMs, and even to improve the architecture designs of MLLMs. 




\section{Related Works}

Exploiting the local and global semantic representations is commonly employed in the processing of image data~\cite{bian2017fusing,lv-etal-2019-an-end,chen-etal-2021-rethinking,zhao2022fuse}. By adjusting the receptive field size of CNN layers, the model can capture information at various granularities~\cite{simonyan2015deep,dai2023rfcn}. However, in the MLLMs, the structure of each Transformer layer can usually be similar or the same to others. Therefore, we wonder how MLLMs represent the local and global information, especially when the inputs are sequences of visual tokens but not matrices of pixels. 

Previous studies~\cite{chi-etal-2020-finding,vulic-etal-2020-probing} in pre-trained language models~(PLMs, i.e., BERT) employ probing tasks to investigate which layer in the model make the most effort to encode lexical, syntactic, or semantic information. These works reveal that the lower layers in BERT can encode lexical information, while the upper ones tend to encode syntactic and semantic information. In this work, we follow previous works and employ multimodal probing tasks to study the granularity of information represented by each layer in decoder-only MLLMs. 

We also note that there are other methods available for studying the representing mechanisms of LLMs. For example, \citet{Sajjad-eteal-2023} propose removing specific layers of PLMs to investigate their effects by comparing the performance of the models before and after removal. Previous works~\cite{kovaleva-etal-2019-revealing,rogers-etal-tacl} also use neuron-wise examinations and visualization methods to provide detailed analyses. Although these methods are mainly implemented on 
encoder-only PLMs, we believe that they may provide insights for future research on the interpretability of MLLMs.

\section{Global Multimodal Representation}

We first aim to investigate how 
each layer can encode the global cross-modal semantic information. Motivated by natural language inference, where the 
alignment between global meanings of two sentences 
plays a vital role~\cite{maccartney-etal-2008-phrase,tay-etal-2018-compare}, we design an image-text entailment task whose goal is to decide whether a caption can entail a given image or not. 

We thus build a dataset based on 
MS COCO~\cite{lin-mscoco-2014}, which contains more than 200K labeled images and five captions for each image. Formally, we denote the images as $\mathcal{M}=\left\{m_i\right\vert i=0,\,1,\,\cdots\}$ and the five human-written caption texts of $m_i$ as $\mathcal{T}_i=\left\{t_{i,k}|k=0,\,\cdots,\,4\right\}$.

We define this image-text semantic entailment task as a binary classification task. 
For each image $m_i\in\mathcal{M}$, we select all its captions $\mathcal{T}_i$ to construct positive image-text pairs. Furthermore, we also use the image $m_i$ and captions sampled from $\bigcup_{j\neq i}\mathcal{T}_j$ to form negative examples. For each positive pair $\left<m_i, \,t_{i,k}\right>$, we randomly sample 5,000 captions from $\bigcup_{j\neq i}\mathcal{T}_j$. To ensure a balanced number of positive and negative examples, we select one negative caption that has the highest similarity\footnote{The similarity is calculated by model \texttt{all-mpnet-base-v2}~(\url{https://huggingface.co/sentence-transformers/all-mpnet-base-v2}).} with $t_{i,k}$ as the negative sample.


Following previous probing studies~\cite{hupkes-probing-2019,jawahar-etal-2019-bert}, we freeze all parameters of the multimodal large language model~(MLLM). We use the following prompt to combine the image and caption pairs as input: \underline{\texttt{[Image]}~\textit{This image describes "}\texttt{[Caption]}\textit{". Is it right? Answer:}}. We then extract the hidden-state features generated by each layer of MLLM, and take the vectors corresponding to the last tokens as representations of the whole inputs. For the $L$-th layer, whose feature vector can be denoted as $\mathcal{H}_L$, we train a binary classifier $f_L: \mathcal{H}_L\mapsto \left\{0,\,1\right\}$. In this paper, we employ single-layer linear classifiers for experiments and use Adam~\cite{adam} as the optimizer. 

\begin{figure}[t] 
\centering 
\resizebox{1\columnwidth}{!}{  
    \input{figure/tikzpic1.tex}
}
\caption{Performance on the image-text entailment task when using the representations at each layer.} 
\label{fig:nli_results}  
\end{figure}
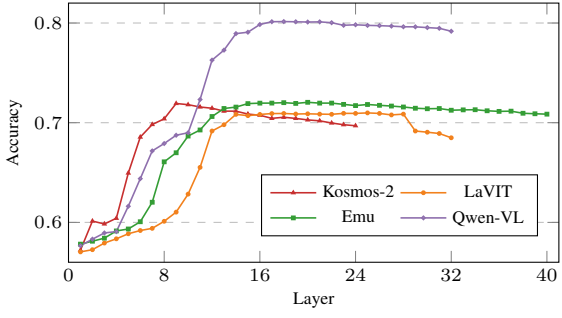

We examine several popular decoder-only MLLMs, including Kosmos-2~\cite{peng2023kosmos2}, LaVIT~\cite{jin2023unified}, Emu\cite{sun2023generative} and Qwen-VL~\cite{Qwen-VL}, with parameter scales ranging from 7B to 14B. We employ \textit{accuracy} to measure the extent to which the representation vectors can encode information for the image-text entailment task. Experimental results are illustrated in Figure~\ref{fig:nli_results}. From the results, we find representation vectors of the topmost layers do not yield optimal performance. For instance, in Kosmos-2, a model consisting of 24 Transformer layers, we find that its representation vectors generated by the 9th layer demonstrate the best performance in the image-text entailment task. Similarly, the 23rd layer in LaVIT, the 20th layer in Emu, and the 18th layer in Qwen-VL achieve the best performance, all of which are not the topmost layers in their corresponding models. As the depth of layers increases to the topmost, all models' abilities to encode global multimodal information exhibit a diminishing trend.

In light of previous research, which demonstrates the upper layers in BERT can possess the strongest ability to represent global semantic information~\cite{jawahar-etal-2019-bert,koto-etal-2021-discourse}, we intuitively hypothesize that the same phenomenon may appear in MLLMs. 
  However, our experimental results display a deviation from the conclusions of prior works on encoder-only PLMs. 

Revisiting the pre-training process of decoder-only MLLMs, we find there is a gap between their pre-training objective and the ability to encode global semantic information. Since models learn how to generate the sequence token by token, the representation vectors encoded by upper layers may inherently focus more on information related to the local token which will be generated next, rather than all the context tokens. 
For MLLMs that have been pre-trained but without being fine-tuned on downstream tasks, their predicted tokens in zero-shot scenarios may not always perform well in addressing complex tasks that need global information. Hence, focusing on encoding the local semantic features of such tokens does not contribute to addressing the image-text entailment task. This may be the reason why representation vectors of intermediate layers outperform the upper layers. 

\section{Local Multimodal Representation}
\label{sec:obj_det}

To investigate whether the upper layers encode more local information about the token to be generated than the lower layers, we employ the MS COCO dataset again and conduct an object recognition task. MS COCO comprises annotations for 80 distinct categories. For an image $m_i$, its annotated object category list can be denoted as $\left\{\mathcal{O}_{k}\vert k\in\mathbb{I}\right\}$, where $\mathbb{I}\subseteq\left\{0,\,1,\,\cdots,\,79\right\}$ is an indicator set denoting the categories of objects present in image $m_i$. In this work, we regard the recognition task for different object categories as 80 separate binary classification tasks, in which the model needs to predict whether the input image contains a specific type of object. 

We first study whether the feature vectors encoded by each layer can be used to predict a specific object category, when we provide sufficient cues of other categories. 
It is intuitive that if there are $n$ types of objects in an image and we provide $(n-1)$ categories of them in the text input, a well pre-trained MLLM should then output the $n$-th remaining category. Thus, we extract the vision-language features by following prompt: \underline{\texttt{[Image]} \textit{This image contains the following types }} \underline{\textit{of objects:} \texttt{[Obj\_1]}\textit{,} \texttt{[Obj\_2]}\textit{, ..., }\texttt{[Obj\_n-1]}\textit{,}}. Similar to the entailment task, we also freeze the parameters of MLLMs and collect representation vectors of the last token as features of the whole input sequence. We denote the representation vectors of layer $L$ as $\mathcal{H}^{\textnormal{WithCat}}_{L}$. It is important to note that, to prevent data leakage, when training and evaluating the probing model for the category $c$, the input object list for image $m_i$ should be $\left\{\mathcal{O}_k\vert k\in\mathbb{I} \wedge k\neq c\right\}$. To mitigate the potential impact arising from the order of input object categories, we shuffle the object lists during both training and evaluation.

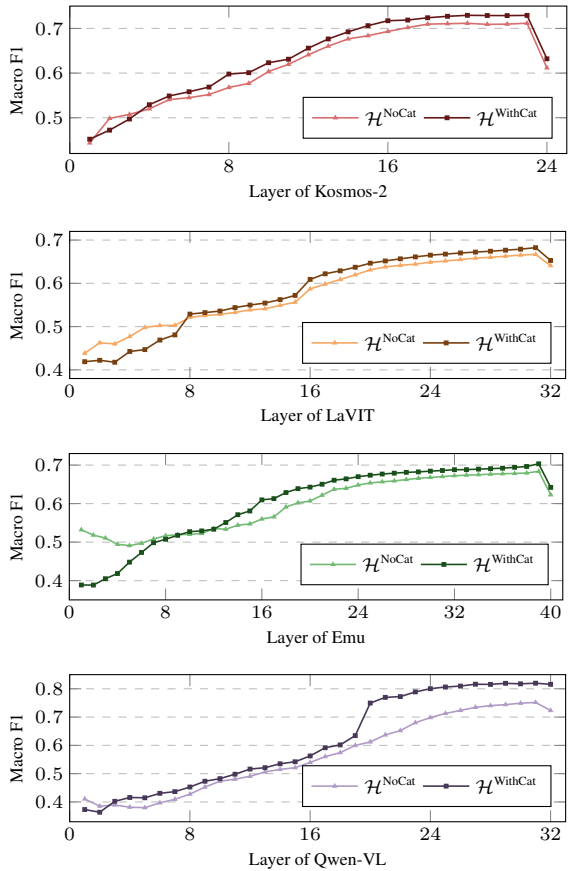
\begin{figure}[t] 
\centering 
\begin{subfigure}{1\columnwidth}
\resizebox{1\columnwidth}{!}{  
    \input{figure/tikzpic2-1.tex}
}
\end{subfigure}
\vskip 0.1in

\begin{subfigure}{1\columnwidth}
\resizebox{1\columnwidth}{!}{  
    \input{figure/tikzpic2-2.tex}
}
\end{subfigure}
\vskip 0.1in

\begin{subfigure}{1\columnwidth}
\resizebox{1\columnwidth}{!}{  
    \input{figure/tikzpic2-3.tex}
}
\end{subfigure}
\vskip 0.1in

\begin{subfigure}{1\columnwidth}
\resizebox{1\columnwidth}{!}{  
    \input{figure/tikzpic2-4.tex}
}
\end{subfigure}
\caption{Performance on the object recognition task when using the representations at each layer of different MLLMs.} 
\label{fig:decode_exp}  
\end{figure}

In the probing study, 
as we look at higher layers,
the improvement of model performance might be attributed to the expansion of the model's parameter scale, resulting in enhanced representing ability. To eliminate the influence of scale expansion on performance, we formulate another set of experiments to perform object recognition without any category cues, serving as the baseline. We use the following prompt: \underline{\texttt{[Image]} \textit{This image contains the following types }} \underline{\textit{of objects:}}. The vector set of layer $L$ extracted by this prompt is denoted as $\mathcal{H}^{\textnormal{NoCat}}_{L}$.

Similar to the entailment task, we also examine the four large-scale multimodal models, including Kosmos-2, LaVIT, Emu, and Qwen-VL. Due to the imbalance in the ratio of positive to negative examples in the object recognition tasks, we employ the Macro Average F1 score across all categories as our evaluation metric. The experimental results are illustrated in Figure \ref{fig:decode_exp}. From the results, we first find that, across the upper layers of all four MLLMs, the probing models trained on $\mathcal{H}^{\textnormal{WithCat}}$ outperform the ones trained on $\mathcal{H}^{\textnormal{NoCat}}$. However, in the lowermost layers, providing several categories as input can hurt the probing model's performance. The results probably indicate that the upper layers, those closer to the token probability prediction layer, tend to encode more local features of the tokens to be decoded, rather than global semantic information.

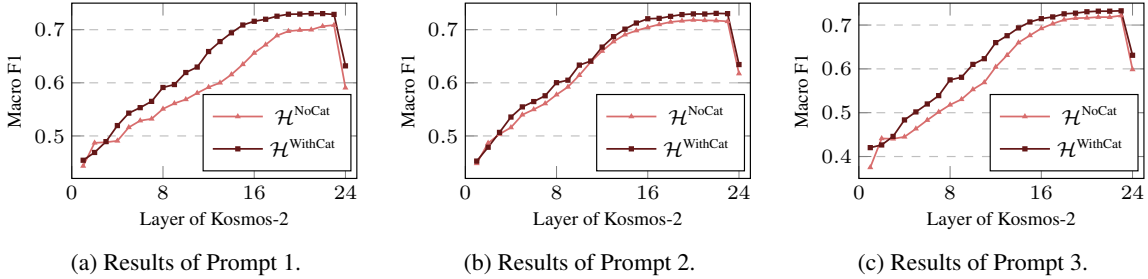
\begin{figure*}[t] 
\centering 
\begin{subfigure}{0.66\columnwidth}
\resizebox{1\columnwidth}{!}{  
    \input{figure/tikzpic3-1.tex}
}
\caption{Results of Prompt 1.}
\end{subfigure}
\begin{subfigure}{0.66\columnwidth}
\resizebox{1\columnwidth}{!}{  
    \input{figure/tikzpic3-2.tex}
}
\caption{Results of Prompt 2.}
\end{subfigure}
\begin{subfigure}{0.66\columnwidth}
\resizebox{1\columnwidth}{!}{  
    \input{figure/tikzpic3-3.tex}
}
\caption{Results of Prompt 3.}
\end{subfigure}
\caption{Performance on object recognition of the representations extracted with various prompts.} 
\label{fig:robust_test}  
\end{figure*}

To further validate the hypothesis, it is necessary to examine whether given a subset of object categories as input, the model indeed produces tokens that are relevant to the remaining categories present in the image. We take the \texttt{person} category and the Kosmos-2 model as a case study. We randomly select 10,000 examples from the test set, with 5,409 of them containing the \texttt{person} objects~(positive examples), while the remaining images do not~(negative examples). We employ the two prompts to extract $\mathcal{H}^{\textnormal{NoCat}}$ and $\mathcal{H}^{\textnormal{WithCat}}$ as input, capturing the first new tokens generated by Kosmos-2. We then examine the frequency distributions of the generated tokens on positive and negative examples separately. In Table \ref{tab:token_stat}, we list the statistical results for the top 10 most frequently generated tokens for each setting. 

\begin{table}[t]
\centering
\begin{tiny}
\begin{tabular}{p{0.6cm}r|p{0.6cm}r|p{0.6cm}r|p{0.6cm}r}
\hline
\multicolumn{4}{c|}{\textbf{Prompt without Categories}} & \multicolumn{4}{c}{\textbf{Prompt with Categories}} \\
\hline
\multicolumn{2}{c|}{\textbf{Pos. Set}} & \multicolumn{2}{c|}{\textbf{Neg. Set}} & \multicolumn{2}{c|}{\textbf{Pos. Set}} & \multicolumn{2}{c}{\textbf{Neg. Set}} \\
\textbf{Token} & \textbf{Freq.} & \textbf{Token} & \textbf{Freq.} & \textbf{Token} & \textbf{Freq.} & \textbf{Token} & \textbf{Freq.}\\
\hline
A & .9662 & A & .9532 & \darkred{man} & \darkred{.1063} & and & .1250\\
a & .0079 & a & .0129 & \darkred{people} & \darkred{.1017} & grass & .0416\\
black & .0063 & \darkblue{zebra} & \darkblue{.0102} & \darkred{woman} & \darkred{.0636} & building & .0281\\ 
two & .0035 & Gir & .0059 & and & .0514 & street & .0279\\
\darkblue{tennis} & \darkblue{.0022} & two & .0044 & tennis & .0431 & mirror & .0229\\
\darkblue{snowboarder} & \darkblue{.0020} & \darkblue{elephant} & \darkblue{.0017} & baseball & .0373 & tree & .0227\\
an & .0018 & an & .0009 & \darkred{person} & \darkred{.0294} & water & .0187\\
\darkblue{baseball} & \darkblue{.0015} & brown & .0009 & beach & .0242 & window & .0170\\
Two & .0013 & Two & .0007 & \darkred{boy} & \darkred{.0238} & animal & .0161\\
\darkblue{skateboard} & \darkblue{.0013} & \darkblue{bananas} & \darkblue{.0007} & skateboard & .0213 & plate & .0157\\
\texttt{OTHERS} & .0059 & \texttt{OTHERS} & .0087 & \texttt{OTHERS} & .4979 & \texttt{OTHERS} & .6643\\
\hline
\end{tabular}
\end{tiny}
\caption{\label{tab:token_stat} Frequency of the top 10 frequently generated tokens.}
\end{table}

We can find when employing the prompt that includes all object categories except \texttt{person}, there is a significant difference in the distributions of the first tokens generated by the model for positive and negative examples. We note that, in the case of positive examples, \textbf{5 out of the top 10} most frequently generated tokens have meanings associated with \textit{person}~(red-colored), while all of the 10 tokens of negative examples lack semantic relevance to \textit{person}.

Nevertheless, when using the prompt without providing any categories, the model generates \textit{"A"} or \textit{"a"} with a frequency exceeding 96\%, both in the positive and negative example sets. We also find among the top 10 frequent tokens, several of them convey meanings corresponding to the object categories other than \texttt{person}, such as \textit{tennis}, \textit{snowboarder}, \textit{zebra}, and other tokens that are colored by blue. It indicates that the representation vectors may randomly encode one category of the objects appearing in the image.

By comparing the results of positive and negative sets generated by the two prompts separately, we can infer that the topmost layer of a MLLM can be effective in representing the local semantic features of the token to be decoded.

Furthermore, we note the model's performance continuously improves from the lowermost to the second-to-last layer, while it significantly declines in the topmost layer. We revisit the frequency distributions of the first generated tokens. We can find there is an overlap in the tokens generated by the model for positive and negative examples, such as \textit{"A"}, \textit{"a"}, \textit{"and"}, \textit{"skateboard"}, and etc. These overlapping tokens may indicate the model produces indistinguishable representation vectors, which negatively affect the performance of probing models. We conjecture that, compared to preceding layers, representation vectors in the topmost layer of a MLLM may lose certain global semantic information but shift their focus towards specific tokens to be predicted, although these tokens may not have relevant meanings to \textit{person}. This could also be the reason why the intermediate layers, rather than the topmost layers, perform better in the image-text entailment tasks.

\section{Results of More Prompts}

Through a pair of cross-modal prompts, we find in the decoder-only MLLMs, the deficiency of upper layers in encoding global semantic information may arise from that such layers focus excessively on the local information of one token. In order to examine the robustness of our findings, we also conduct experiments with different prompts to probe the model's ability to perform object recognition tasks. The employed prompts are listed in Table \ref{tab:prompts}. These prompts use diverse forms of expressions and possess varying lengths. 

We replace the prompts in Section~\ref{sec:obj_det} with them and implement experiments based on the same settings. We take Kosmos-2 as an instance, and the results based on each prompt are shown in Figure~\ref{fig:robust_test}. Comparing the results in Figure \ref{fig:decode_exp} and Figure \ref{fig:robust_test}, we find despite using different prompts, the Kosmos-2 model performs consistently across all groups of experiments. It indicates that our findings can be prompt-agnostic.
\begin{table}
\centering
\begin{scriptsize}
\begin{tabular}{p{7.2cm}}
\hline
\textbf{Prompt 1}: \texttt{[Image]} \textit{What types of objects are there here? Please list them:} \texttt{[Obj\_1]}\textit{,} \texttt{[Obj\_2]}\textit{, ..., }\texttt{[Obj\_k]}\textit{,}\\
\hline
\textbf{Prompt 2}: \texttt{[Image]} \textit{Objects in this picture are:} \texttt{[Obj\_1]}\textit{,}\\ \texttt{[Obj\_2]}\textit{, ..., }\texttt{[Obj\_k]}\textit{,}\\
\hline
\textbf{Prompt 3}: \texttt{[Image]} \textit{There can be several types of objects in this image, including up to eighty kinds of objects. These objects can be any color, including red, green, blue, orange, yellow, purple, pink, and etc. Some of these objects can be very huge, while others can be very small. In the meantime, there are also many objects which can be overlapping with others. Please look carefully at the image for any detailed information. Now, you can write which type of objects you can find in the image:} \texttt{[Obj\_1]}\textit{,} \texttt{[Obj\_2]}\textit{, ..., }\texttt{[Obj\_k]}\textit{,}\\
\hline
\end{tabular}
\end{scriptsize}
\caption{\label{tab:prompts} Variant prompts to extract representation vectors for object recognition.}
\end{table}

\section{Conclusion}

In this paper, we investigate how the decoder-only MLLMs represent the global and local cross-modal semantic information, through prompt-based probing study. We experiment with four open-source models, extracting representation vectors using various prompts. Our findings remain consistent across diverse models and prompts, which indicates \textbf{the upper layers in MLLMs focus too much on} the semantic features of \textbf{the next token} to be generated. It may result in a loss of global information in the upper layers. Our findings shed light on understanding the potential mechanism of MLLMs to represent global and local features. We hope this paper can inspire our community to delve into more effective pre-training mechanisms for MLLMs.

\section{Acknowledgement}

This work is supported in part by NSFC~(62161160339) and Kuaishou. We would like to thank the anonymous reviewers for their helpful comments and suggestions.

\bibliography{lrec-coling2024-example}
\bibliographystyle{acl_natbib}

\end{document}

%% file: figure/tikzpic1.tex
\begin{tikzpicture} 
    \scalefont{0.6} 
    \begin{axis}[
    sharp plot, 
    xmode=normal,
    xlabel=Layer, 
    ylabel=Accuracy, 
    width=8cm, height=5cm,  
    xmin=0,xmax=41,  
    ymin=0.56, ymax=0.82,  
    xtick={0,8,16,24,32,40}, 
    ytick={0.60,0.70,0.80}, 
    xlabel near ticks, 
    ylabel near ticks, 
    ymajorgrids=true, 
    grid style=dashed, 
    legend style={at={(0.675,0.1)},anchor=south}, 
    legend columns=2, 
    ]
    
    \addplot+[semithick,mark=triangle*,mark options={scale=0.3}, color=color1] plot coordinates { 
        (1, 0.5721624048871475) (2, 0.60130567370987) (3, 0.598414045614693) (4, 0.6041479566157095) (5, 0.6494863165789967) (6, 0.6855527154657692) (7, 0.6981850839361671) (8, 0.7038795187857135) (9, 0.7192752178590109) (10, 0.7180810642770437) (11, 0.7157421023024465) (12, 0.714513407087943) (13, 0.7116168444738322) (14, 0.7115674992844948) (15, 0.7086561331135828) (16, 0.7074669140505492) (17, 0.704363101641221) (18, 0.7054634993634471) (19, 0.7042989528950823) (20, 0.7027988591392225) (21, 0.7018761040986115) (22, 0.6996901122109606) (23, 0.6979827686598834) (24, 0.6969514542027297)
    };
    \addlegendentry{Kosmos-2}
    
    \addplot+[semithick,mark=*,mark options={scale=0.3}, color=color2] plot coordinates {
        (1, 0.5704007816277991) (2, 0.5725966425533174) (3, 0.5792681121517463) (4, 0.5835216674726381) (5, 0.5885894184175985) (6, 0.5918363318760054) (7, 0.594007520206855) (8, 0.6011082929525201) (9, 0.610296367207161) (10, 0.6283567065046829) (11, 0.6551906204664107) (12, 0.6916863225004194) (13, 0.6979728996220158) (14, 0.7083649964964915) (15, 0.7070622834979818) (16, 0.7082811096746178) (17, 0.7091594540448252) (18, 0.7092630789424339) (19, 0.7087153473407878) (20, 0.7089275316549389) (21, 0.7086067879242453) (22, 0.7084143416858291) (23, 0.7093716383589764) (24, 0.7093272276885726) (25, 0.7098354831387488) (26, 0.7093025550939039) (27, 0.7078073958569779) (28, 0.7085969188863778) (29, 0.691656715386817) (30, 0.6904674963237833) (31, 0.6891647833252736) (32, 0.6848766863718456)
    };
    \addlegendentry{LaVIT} 
    
    \addplot+[semithick,mark=square*,mark options={scale=0.3}, color=color3] plot coordinates {
        (1, 0.5781775834673878) (2, 0.5811185567519023) (3, 0.584173023971893) (4, 0.591411963247703) (5, 0.5933759017833351) (6, 0.6006444481727476) (7, 0.6202098157450631) (8, 0.6607567578236798) (9, 0.6698461416996457) (10, 0.6865988334797241) (11, 0.692697898881838) (12, 0.7062135462413769) (13, 0.7142617466223218) (14, 0.715653280961639) (15, 0.7191666584424684) (16, 0.7196156996654396) (17, 0.7196699793737109) (18, 0.720099282520947) (19, 0.7193541701619509) (20, 0.7204545678841769) (21, 0.7196453067790421) (22, 0.7196699793737109) (23, 0.7183820699320024) (24, 0.7171632437553663) (25, 0.7182636414775924) (26, 0.7174494458535238) (27, 0.7167043334945277) (28, 0.715751971340314) (29, 0.714513407087943) (30, 0.7140100861567006) (31, 0.7142124014329843) (32, 0.7125050578819071) (33, 0.7128751468019383) (34, 0.713018247851017) (35, 0.7119622607991947) (36, 0.7113553149703435) (37, 0.7115724338034285) (38, 0.7094752632565852) (39, 0.7089719423253427) (40, 0.7085771808106428)
    };
    \addlegendentry{Emu} 

    \addplot+[semithick,mark=halfsquare*,mark options={scale=0.3}, color=color4] plot coordinates {
        (1, 0.5765936028896543) (2, 0.5831466440336731) (3, 0.5893986795227333) (4, 0.5907704757863156) (5, 0.6161684447383224) (6, 0.6438510959566552) (7, 0.6716916517808679) (8, 0.6791032992193591) (9, 0.6874130291037926) (10, 0.6899148302032035) (11, 0.7232623091574802) (12, 0.7627828712978771) (13, 0.7727950102144542) (14, 0.7893651247939838) (15, 0.7907714626901023) (16, 0.7985186574160885) (17, 0.8013066606136567) (18, 0.8014448271438017) (19, 0.801242511867518) (20, 0.8009908514018969) (21, 0.801143821488843) (22, 0.8003148223079732) (23, 0.7977192653488211) (24, 0.7982571279125998) (25, 0.79762550948908) (26, 0.7973639799855912) (27, 0.7968705280922163) (28, 0.7962339751497627) (29, 0.7961698264036239) (30, 0.7954148450067603) (31, 0.7947092087992341) (32, 0.7917484974389847)
    };
    \addlegendentry{Qwen-VL} 

    \end{axis}
\end{tikzpicture}

%% file: figure/tikzpic2-1.tex
\begin{tikzpicture} 
    \scalefont{0.6} 
    \begin{axis}[
    sharp plot, 
    xmode=normal,
    xlabel=Layer of Kosmos-2, 
    ylabel=Macro F1, 
    width=8cm, height=3.5cm,  
    xmin=0,xmax=25,  
    ymin=0.42, ymax=0.75,  
    xtick={0,8,16,24}, 
    ytick={0.50,0.60,0.70}, 
    xlabel near ticks, 
    ylabel near ticks, 
    ymajorgrids=true, 
    grid style=dashed, 
    legend style={at={(0.72,0.1)},anchor=south}, 
    legend columns=2, 
    ]
    
    \addplot+[semithick,mark=triangle*,mark options={scale=0.3}, color=color1!70!white] plot coordinates { 
        (1, 0.44359715513698433) (2, 0.49858588859918046) (3, 0.507487173980845) (4, 0.5199073138178426) (5, 0.5405748507313384) (6, 0.5449532647033053) (7, 0.551863691991425) (8, 0.5678338658073213) (9, 0.5768178633857384) (10, 0.6036178031357456) (11, 0.6195900239726299) (12, 0.6409975291535789) (13, 0.6603504806808523) (14, 0.6764467451475815) (15, 0.6836203504387053) (16, 0.6932582884364893) (17, 0.7018818980821108) (18, 0.7096184874683833) (19, 0.7105264913790112) (20, 0.7114769887144391) (21, 0.7090559129259527) (22, 0.7096133055923328) (23, 0.7116443857281086) (24, 0.611724394002165) 
    };
    \addlegendentry{$\mathcal{H}^{\textnormal{NoCat}}$}
    
    \addplot+[semithick,mark=square*,mark options={scale=0.3}, color=color1!50!black] plot coordinates {
        (1, 0.4520535352468985) (2, 0.4723005635999809) (3, 0.4972252199271156) (4, 0.5291973707584203) (5, 0.5487669339517268) (6, 0.5582082261227059) (7, 0.5686480779676254) (8, 0.5978431978493725) (9, 0.6008832134348585) (10, 0.6231739024227607) (11, 0.6310692430944406) (12, 0.6558545467237957) (13, 0.6763994284098663) (14, 0.6923506367122247) (15, 0.7061717008034907) (16, 0.717259712390189) (17, 0.7187298364080019) (18, 0.7238645789439728) (19, 0.7271244471134108) (20, 0.7293223738704726) (21, 0.7289099002809231) (22, 0.7287281894995233) (23, 0.7291513029370637) (24, 0.6320075331646309) 
    };
    \addlegendentry{$\mathcal{H}^{\textnormal{WithCat}}$} 

    \end{axis}
\end{tikzpicture}

%% file: figure/tikzpic2-2.tex
\begin{tikzpicture} 
    \scalefont{0.6} 
    \begin{axis}[
    sharp plot, 
    xmode=normal,
    xlabel=Layer of LaVIT, 
    ylabel=Macro F1, 
    width=8cm, height=3.5cm,  
    xmin=0,xmax=33,  
    ymin=0.38, ymax=0.72,  
    xtick={0,8,16,24,32}, 
    ytick={0.40,0.50,0.60,0.70}, 
    xlabel near ticks, 
    ylabel near ticks, 
    ymajorgrids=true, 
    grid style=dashed, 
    legend style={at={(0.72,0.1)},anchor=south}, 
    legend columns=2, 
    ]
    
    \addplot+[semithick,mark=triangle*,mark options={scale=0.3}, color=color2!70!white] plot coordinates { 
        (1, 0.4384700307218197) (2, 0.4625628021583469) (3, 0.459992583340388) (4, 0.47717171571636036) (5, 0.49800529890168665) (6, 0.5024461372508882) (7, 0.502891506639457) (8, 0.5214530441541628) (9, 0.5253457439909024) (10, 0.528113011543272) (11, 0.5327457803564688) (12, 0.5385451000518461) (13, 0.5411091051297899) (14, 0.5490432409541167) (15, 0.5564192827664605) (16, 0.5870629943634131) (17, 0.5981611960256256) (18, 0.6088227675806784) (19, 0.6196513039450482) (20, 0.6310342095930785) (21, 0.6380319610440276) (22, 0.6416790621249466) (23, 0.6443095863000534) (24, 0.6488815321603638) (25, 0.6516081499080997) (26, 0.6548861223197262) (27, 0.6582410820777651) (28, 0.6601837520492934) (29, 0.66267927930263) (30, 0.6652769613431887) (31, 0.6668955931097119) (32, 0.6409760280409731) 
    };
    \addlegendentry{$\mathcal{H}^{\textnormal{NoCat}}$}
    
    \addplot+[semithick,mark=square*,mark options={scale=0.3}, color=color2!50!black] plot coordinates {
        (1, 0.4188807126745201) (2, 0.42211183585844403) (3, 0.41761712124635625) (4, 0.44249031684721274) (5, 0.4468032621833258) (6, 0.4689122077014127) (7, 0.48081928570611243) (8, 0.5288904327513262) (9, 0.5323920010797591) (10, 0.5359301543494992) (11, 0.543927754248253) (12, 0.5497960487423077) (13, 0.5544377732160497) (14, 0.5624001654844719) (15, 0.5720908630136095) (16, 0.6090731538465252) (17, 0.62207674404714) (18, 0.6289353520198463) (19, 0.637160549479887) (20, 0.6465906255525851) (21, 0.6517437437688797) (22, 0.6566367954404291) (23, 0.6611069731774418) (24, 0.6651123396642143) (25, 0.6675569219984847) (26, 0.670158080216369) (27, 0.6722299911224727) (28, 0.674206349777624) (29, 0.6766087377828397) (30, 0.6789665475797131) (31, 0.6824370270720548) (32, 0.6528174920733119) 
    };
    \addlegendentry{$\mathcal{H}^{\textnormal{WithCat}}$} 

    \end{axis}
\end{tikzpicture}

%% file: figure/tikzpic2-3.tex
\begin{tikzpicture} 
    \scalefont{0.6} 
    \begin{axis}[
    sharp plot, 
    xmode=normal,
    xlabel=Layer of Emu, 
    ylabel=Macro F1, 
    width=8cm, height=3.5cm,  
    xmin=0,xmax=41,  
    ymin=0.35, ymax=0.73,  
    xtick={0,8,16,24,32,40}, 
    ytick={0.40,0.50,0.60,0.70}, 
    xlabel near ticks, 
    ylabel near ticks, 
    ymajorgrids=true, 
    grid style=dashed, 
    legend style={at={(0.72,0.1)},anchor=south}, 
    legend columns=2, 
    ]
    
    \addplot+[semithick,mark=triangle*,mark options={scale=0.3}, color=color3!70!white] plot coordinates { 
        (1, 0.5319285972686685) (2, 0.5182048023282195) (3, 0.5105775745745986) (4, 0.4944270854301629) (5, 0.4909975296343389) (6, 0.49704900217252534) (7, 0.5081277443844737) (8, 0.51660330523956) (9, 0.519392016748175) (10, 0.520112141587293) (11, 0.5224611696652219) (12, 0.534980794149849) (13, 0.5335412785272867) (14, 0.5439640976142187) (15, 0.5471432196928194) (16, 0.5600326302810653) (17, 0.5655663768085921) (18, 0.5910530210175529) (19, 0.6019532667545395) (20, 0.6067589916162314) (21, 0.6218392219956477) (22, 0.6372989683131561) (23, 0.6397238888435985) (24, 0.6481594580986302) (25, 0.653705395415099) (26, 0.6565305137352686) (27, 0.658916073115237) (28, 0.6625086379108475) (29, 0.665674492613797) (30, 0.667922137601885) (31, 0.6702021103736431) (32, 0.6724453900995803) (33, 0.6738980173474607) (34, 0.6750028527816044) (35, 0.6764687808878218) (36, 0.6775031103137604) (37, 0.6783235184580937) (38, 0.6794618943785916) (39, 0.6837613068400072) (40, 0.6227899535132693) 
    };
    \addlegendentry{$\mathcal{H}^{\textnormal{NoCat}}$}
    
    \addplot+[semithick,mark=square*,mark options={scale=0.3}, color=color3!50!black] plot coordinates {
        (1, 0.388664325143272) (2, 0.388482057131988) (3, 0.4049780177757437) (4, 0.4184361763280979) (5, 0.44802385421873164) (6, 0.4731564522160288) (7, 0.49845507859085025) (8, 0.5073996924696608) (9, 0.5177046146874437) (10, 0.5271806557612423) (11, 0.5291083423701075) (12, 0.5334388331730263) (13, 0.5508833979362484) (14, 0.5711316923277702) (15, 0.5809056676351793) (16, 0.609660317473046) (17, 0.6128949337757338) (18, 0.6287504235883141) (19, 0.6388205135107611) (20, 0.6427945519449898) (21, 0.6502576609902279) (22, 0.6606909239332199) (23, 0.6642717289710542) (24, 0.6699528300432038) (25, 0.6736892363945146) (26, 0.6767483659549481) (27, 0.6790332368911589) (28, 0.6811673554303562) (29, 0.6823019113998593) (30, 0.6843321951971086) (31, 0.6859654258990313) (32, 0.6880237971959265) (33, 0.6881147771959588) (34, 0.6895449839715447) (35, 0.690668820655749) (36, 0.6919870918888998) (37, 0.6941283818376843) (38, 0.6964004043786771) (39, 0.7033396191195637) (40, 0.6421243476269954) 
    };
    \addlegendentry{$\mathcal{H}^{\textnormal{WithCat}}$} 

    \end{axis}
\end{tikzpicture}

%% file: figure/tikzpic2-4.tex
\begin{tikzpicture} 
    \scalefont{0.6} 
    \begin{axis}[
    sharp plot, 
    xmode=normal,
    xlabel=Layer of Qwen-VL, 
    ylabel=Macro F1, 
    width=8cm, height=3.5cm,  
    xmin=0,xmax=33,  
    ymin=0.33, ymax=0.85,  
    xtick={0,8,16,24,32}, 
    ytick={0.40,0.50,0.60,0.70,0.80}, 
    xlabel near ticks, 
    ylabel near ticks, 
    ymajorgrids=true, 
    grid style=dashed, 
    legend style={at={(0.72,0.1)},anchor=south}, 
    legend columns=2, 
    ]
    
    \addplot+[semithick,mark=triangle*,mark options={scale=0.3}, color=color4!70!white] plot coordinates { 
        (1, 0.41053869436061213) (2, 0.3850275660763639) (3, 0.3892530433592561) (4, 0.3815308547361087) (5, 0.3796539049982817) (6, 0.3967129505157767) (7, 0.4093808525967151) (8, 0.42766862744091405) (9, 0.45263406259167294) (10, 0.47331921845944747) (11, 0.48052111397537073) (12, 0.4907910757059133) (13, 0.5060259658382525) (14, 0.5150477843621893) (15, 0.5211444674093326) (16, 0.5395302069193397) (17, 0.5601763432483836) (18, 0.5742914796273142) (19, 0.5996114328560355) (20, 0.6122273604681618) (21, 0.6369797491193991) (22, 0.6521483671901418) (23, 0.680146522707908) (24, 0.6981823007720143) (25, 0.7129233495270695) (26, 0.7234318881882521) (27, 0.7345625415230358) (28, 0.7403899279794538) (29, 0.7440428994138162) (30, 0.7490558452252547) (31, 0.751768879252736) (32, 0.7233498844122888)
    };
    \addlegendentry{$\mathcal{H}^{\textnormal{NoCat}}$}
    
    \addplot+[semithick,mark=square*,mark options={scale=0.3}, color=color4!50!black] plot coordinates {
        (1, 0.37315958622498724) (2, 0.3636381504359308) (3, 0.4022505479879704) (4, 0.4157617380789936) (5, 0.4146980404188315) (6, 0.4304782278152645) (7, 0.43644800809986545) (8, 0.4527468337250194) (9, 0.4731582609007162) (10, 0.48240381944953975) (11, 0.49802167305570366) (12, 0.5162037788057569) (13, 0.5211052451943177) (14, 0.5348514880804487) (15, 0.5421133879630291) (16, 0.5626333734166332) (17, 0.5916934082563621) (18, 0.6017029685534456) (19, 0.6344673637606869) (20, 0.7495722062372632) (21, 0.7697102185149236) (22, 0.7724135822888587) (23, 0.7889529335135901) (24, 0.8004139041223148) (25, 0.8062820172849949) (26, 0.8096326688279942) (27, 0.816170748065806) (28, 0.8155637579182722) (29, 0.8193716892394454) (30, 0.8175549545326994) (31, 0.8197777050841648) (32, 0.8158487095824599) 
    };
    \addlegendentry{$\mathcal{H}^{\textnormal{WithCat}}$} 

    \end{axis}
\end{tikzpicture}

%% file: figure/tikzpic3-1.tex
\begin{tikzpicture} 
    \scalefont{0.6} 
    \begin{axis}[
    sharp plot, 
    xmode=normal,
    xlabel=Layer of Kosmos-2, 
    ylabel=Macro F1, 
    width=5cm, height=3.7cm,  
    xmin=0,xmax=25,  
    ymin=0.42, ymax=0.75,  
    xtick={0,8,16,24}, 
    ytick={0.50,0.60,0.70}, 
    xlabel near ticks, 
    ylabel near ticks, 
    ymajorgrids=true, 
    grid style=dashed, 
    legend style={at={(0.72,0.04)},anchor=south}, 
    legend columns=1, 
    ]
    
    \addplot+[semithick,mark=triangle*,mark options={scale=0.3}, color=color1!70!white] plot coordinates { 
        (1, 0.4434232666970879) (2, 0.4869743495131849) (3, 0.4881273876696711) (4, 0.49073090141509884) (5, 0.5161150069286501) (6, 0.5287895495109532) (7, 0.5321512339915976) (8, 0.550893218893395) (9, 0.5612361485762878) (10, 0.5687607537452892) (11, 0.5806816269432408) (12, 0.5917409571109218) (13, 0.6000756827548236) (14, 0.6155896775555856) (15, 0.6347091739342272) (16, 0.6561485401407477) (17, 0.6716423535989853) (18, 0.6888933897331513) (19, 0.6971683419720337) (20, 0.6992924260131137) (21, 0.6998965238778062) (22, 0.7064558463649855) (23, 0.7086919454400613) (24, 0.5906241390538876)  
    };
    \addlegendentry{$\mathcal{H}^{\textnormal{NoCat}}$}
    
    \addplot+[semithick,mark=square*,mark options={scale=0.3}, color=color1!50!black] plot coordinates {
        (1, 0.4542526138226621) (2, 0.4688977122888658) (3, 0.48930639177695634) (4, 0.5195118315716056) (5, 0.5429607140236861) (6, 0.5532503234383566) (7, 0.5650569817912977) (8, 0.5912194398801001) (9, 0.5967456885774469) (10, 0.6193847776845438) (11, 0.6296392242577771) (12, 0.6589007276236535) (13, 0.6776281267023596) (14, 0.6941177056470325) (15, 0.7087907154730813) (16, 0.7158099166344943) (17, 0.7196849449575732) (18, 0.7253946552283425) (19, 0.7291314374899669) (20, 0.7293711446259588) (21, 0.7302688038129903) (22, 0.7302593208429539) (23, 0.7287879938834837) (24, 0.6319252435735155) 
    };
    \addlegendentry{$\mathcal{H}^{\textnormal{WithCat}}$} 

    \end{axis}
\end{tikzpicture}

%% file: figure/tikzpic3-2.tex
\begin{tikzpicture} 
    \scalefont{0.6} 
    \begin{axis}[
    sharp plot, 
    xmode=normal,
    xlabel=Layer of Kosmos-2, 
    ylabel=Macro F1, 
    width=5cm, height=3.7cm,  
    xmin=0,xmax=25,  
    ymin=0.42, ymax=0.75,  
    xtick={0,8,16,24}, 
    ytick={0.50,0.60,0.70}, 
    xlabel near ticks, 
    ylabel near ticks, 
    ymajorgrids=true, 
    grid style=dashed, 
    legend style={at={(0.72,0.04)},anchor=south}, 
    legend columns=1, 
    ]
    
    \addplot+[semithick,mark=triangle*,mark options={scale=0.3}, color=color1!70!white] plot coordinates { 
        (1, 0.44832278499656225) (2, 0.4865840850205497) (3, 0.5033881308172734) (4, 0.515863598497581) (5, 0.5398374468760065) (6, 0.549874078260085) (7, 0.5609938835510804) (8, 0.5776632772659069) (9, 0.5922780546127057) (10, 0.6143351375069446) (11, 0.6391564780294123) (12, 0.6598185762079166) (13, 0.6775849679622113) (14, 0.690657316597417) (15, 0.6982946946207136) (16, 0.704400794634856) (17, 0.7098923846499621) (18, 0.7145342515269905) (19, 0.7166328861568995) (20, 0.718287224426721) (21, 0.7175738880638444) (22, 0.7170731277366391) (23, 0.7156903000405884) (24, 0.6173374343496473) 
    };
    \addlegendentry{$\mathcal{H}^{\textnormal{NoCat}}$}
    
    \addplot+[semithick,mark=square*,mark options={scale=0.3}, color=color1!50!black] plot coordinates {
        (1, 0.4526774712752551) (2, 0.4787393435509452) (3, 0.5066768186877431) (4, 0.5355742081220959) (5, 0.5546764089921946) (6, 0.5647124309457112) (7, 0.5756185715102963) (8, 0.600257689208024) (9, 0.6049367445873336) (10, 0.6333339668670255) (11, 0.6408903416339562) (12, 0.6672307879953381) (13, 0.6869455586579285) (14, 0.7010452997021915) (15, 0.7125401839719532) (16, 0.7206236954458365) (17, 0.7213507022778869) (18, 0.7251528289260056) (19, 0.7284818414933057) (20, 0.7297120107305648) (21, 0.7295158208780205) (22, 0.7307252154866186) (23, 0.7302266633685732) (24, 0.634376843255744) 
    };
    \addlegendentry{$\mathcal{H}^{\textnormal{WithCat}}$} 

    \end{axis}
\end{tikzpicture}

%% file: figure/tikzpic3-3.tex
\begin{tikzpicture} 
    \scalefont{0.6} 
    \begin{axis}[
    sharp plot, 
    xmode=normal,
    xlabel=Layer of Kosmos-2, 
    ylabel=Macro F1, 
    width=5cm, height=3.7cm,  
    xmin=0,xmax=25,  
    ymin=0.35, ymax=0.75,  
    xtick={0,8,16,24}, 
    ytick={0.40,0.50,0.60,0.70}, 
    xlabel near ticks, 
    ylabel near ticks, 
    ymajorgrids=true, 
    grid style=dashed, 
    legend style={at={(0.72,0.04)},anchor=south}, 
    legend columns=1, 
    ]
    
    \addplot+[semithick,mark=triangle*,mark options={scale=0.3}, color=color1!70!white] plot coordinates { 
        (1, 0.3751348662106221) (2, 0.4412423675053315) (3, 0.44042040825268874) (4, 0.44489431799805823) (5, 0.46342643060615485) (6, 0.48313573076636407) (7, 0.5015207998648838) (8, 0.5179689738737789) (9, 0.530496544097269) (10, 0.5530109127992213) (11, 0.5689987437020391) (12, 0.6042882442904256) (13, 0.6311468848414691) (14, 0.659880611285448) (15, 0.676300870319623) (16, 0.6922366848574317) (17, 0.7029356692792815) (18, 0.7120714927144242) (19, 0.7157740070358376) (20, 0.7163676373643337) (21, 0.7181639332951187) (22, 0.7179935873208745) (23, 0.7216144165227271) (24, 0.5986096384892036)
    };
    \addlegendentry{$\mathcal{H}^{\textnormal{NoCat}}$}
    
    \addplot+[semithick,mark=square*,mark options={scale=0.3}, color=color1!50!black] plot coordinates {
        (1, 0.42030060373095013) (2, 0.4264061442761518) (3, 0.4458769895239345) (4, 0.48348914049075464) (5, 0.5017882515510491) (6, 0.5199307723403237) (7, 0.5389682796983575) (8, 0.5748679901576528) (9, 0.5808050470021657) (10, 0.6102495839860713) (11, 0.6232551325201234) (12, 0.6599377772773801) (13, 0.675681465091009) (14, 0.6934868636168924) (15, 0.7070002583504429) (16, 0.7145867923152905) (17, 0.7185591427584953) (18, 0.7255917106534944) (19, 0.7271404700592616) (20, 0.7303282637284749) (21, 0.731643484034967) (22, 0.7316295117229556) (23, 0.7324871306921023) (24, 0.6310394514273425) 
    };
    \addlegendentry{$\mathcal{H}^{\textnormal{WithCat}}$} 

    \end{axis}
\end{tikzpicture}